\documentclass[journal]{IEEEtran}
\ifCLASSINFOpdf
\else
\fi

\usepackage{graphicx}
\usepackage{xcolor,colortbl}
\usepackage{multirow}
\usepackage{amsmath}

\usepackage{scalerel}
\usepackage{tikz}
\usetikzlibrary{svg.path}

\definecolor{orcidlogocol}{HTML}{A6CE39}
\tikzset{
  orcidlogo/.pic={
    \fill[orcidlogocol] svg{M256,128c0,70.7-57.3,128-128,128C57.3,256,0,198.7,0,128C0,57.3,57.3,0,128,0C198.7,0,256,57.3,256,128z};
    \fill[white] svg{M86.3,186.2H70.9V79.1h15.4v48.4V186.2z}
                 svg{M108.9,79.1h41.6c39.6,0,57,28.3,57,53.6c0,27.5-21.5,53.6-56.8,53.6h-41.8V79.1z M124.3,172.4h24.5c34.9,0,42.9-26.5,42.9-39.7c0-21.5-13.7-39.7-43.7-39.7h-23.7V172.4z}
                 svg{M88.7,56.8c0,5.5-4.5,10.1-10.1,10.1c-5.6,0-10.1-4.6-10.1-10.1c0-5.6,4.5-10.1,10.1-10.1C84.2,46.7,88.7,51.3,88.7,56.8z};
  }
}

\newcommand\orcidicon[1]{\href{https://orcid.org/#1}{\mbox{\scalerel*{
\begin{tikzpicture}[yscale=-1,transform shape]
\pic{orcidlogo};
\end{tikzpicture}
}{|}}}}

\usepackage{hyperref} %<--- Load after everything else

\begin{document}
\title{Seismic Shot Gather Noise Localization Using a Multi-Scale Feature-Fusion-Based Neural Network}

\author{Antonio Jos\'e~G.~Busson~\orcidicon{0000-0001-5394-0707},
        S\'ergio~Colcher~\orcidicon{0000-0002-3476-8718},
        Ruy~Luiz~Milidi\'u~\orcidicon{0000-0002-3423-9998},
        Bruno~Pereira~Dias~\orcidicon{0000-0002-4800-5797}
        and~Andr\'e~Bulc\~ao~\orcidicon{0000-0002-9871-9683}
        
\thanks{A. J. G. Busson, S. Colcher and R. L. Milidi\'u are with the Department of Informatics, Pontifical Catholic  University  of  Rio  de  Janeiro,  Rio  de  Janeiro,  22451-900,  Brazil (e-mail: busson@telemidia.puc-rio.br).}% <-this % stops a space
\thanks{B. Dias and A. Bulc\~ao are with the Leopoldo Am\'erico Miguez de Mello Research and Development Center (Cenpes), Petrobras, Fund\~ao Island, Rio de Janeiro, 21941-970, Brazil.}% <-this % stops a space
\thanks{Manuscript received April 19, 2019; revised September 17, 2019.}}

\maketitle

\begin{abstract}
Deep learning-based models, such as convolutional neural networks, have advanced various segments of computer vision. However, this technology is rarely applied to seismic shot-gather noise localization problem. This letter presents an investigation on the effectiveness of a multi-scale feature-fusion-based network for seismic shot-gather noise localization. Herein, we describe the following: (1) the construction of a real-world dataset of seismic noise localization based on 6,500 seismograms; (2) a multi-scale feature-fusion-based detector that uses the MobileNet combined with the Feature Pyramid Net as the backbone; and (3) the Single Shot multi-box detector for box classification/regression. Additionally, we propose the use of the Focal Loss function that improves the detector's prediction accuracy. The proposed detector achieves an AP@0.5 of 78.67\% in our empirical evaluation.
\end{abstract}


% \fi
\IEEEpeerreviewmaketitle

\section{Introduction}
\label{sec:introduction}

A classic challenge in the field of geophysics involves properly estimating the characteristics of the Earth's subsurface based on measurements acquired by sensors on the surface. Seismic reflection is one of the most widely used methods. It involves generating seismic waves using controlled active sources on the surface (e.g., dynamite explosions in land acquisition or air guns in marine acquisition), and further collecting the reflected data with sensors located above the area~\cite{duarte2014seismic}. The term \textit{shot} refers to a firing by one of these sources. By grouping the seismic signals resulting from the same shot and registered by the sensors into a common-shot domain called the \textit{shot gather} makes it possible to produce an image that represents information about that Earth's subsurface area~\cite{yilmaz2001seismic}.

\begin{figure}
    \centering
    \includegraphics[width=0.35\textwidth]{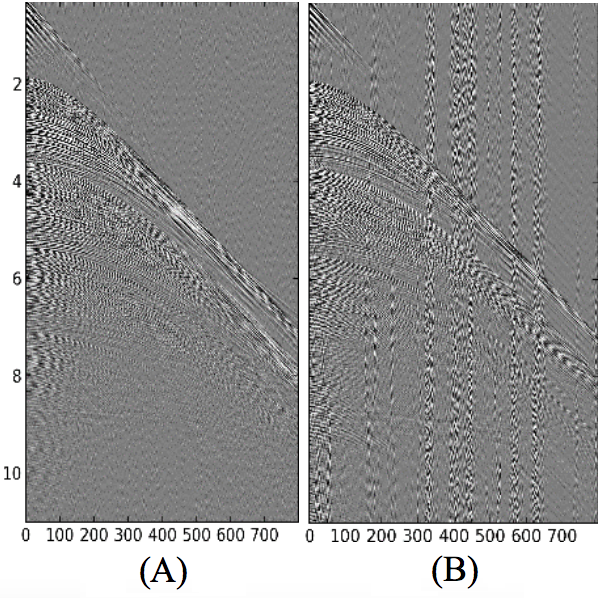}
    \caption{Examples of shot gather images classified by a geophysicist as (A) ``Good" and (B) ``Bad", according to
their noise intensity.}
    \label{fig:seismic_classification}
\end{figure}

\begin{figure}[h]
    \centering
    \includegraphics[width=0.5\textwidth]{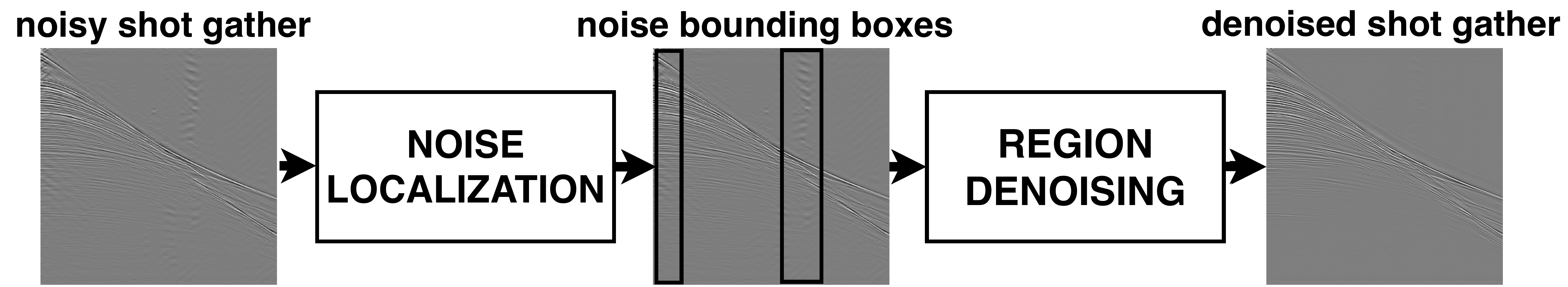}
    \caption{Seismic  \textit{shot gather} region denoising: First, given a noisy shot gather, the \textbf{noise localization} process returns a list of bounding boxes around noise regions. Next, the \textbf{region denoising} process filters the regions delimited by bounding boxes, producing a final denoised seismic  \textit{shot gather}.}
    \label{fig:flow}
\end{figure}

As illustrated in Fig.~\ref{fig:seismic_classification}, seismic \textit{shot gather} data generally contain noise, and the localization and removal of this noise are critical in the early stages of seismic processing. Rather than using various inefficient visual quality control techniques, machine learning detectors can be used to reduce the turnaround time by quickly identifying poor-quality shot gather regions. Next, machine learning-based denoisers~\cite{zhang2017beyond} or standard seismic filters~\cite{elboth2009attenuation} can locally improve the quality of these regions by noise attenuation and/or removal without interfering with the global desired signal data. Fig.~\ref{fig:flow} illustrates this process.

In a \textit{shot gather} image, the abscissa represents the position of the sensor relative to the shot position. According to this, a seductive idea comes to our mind: if the seismic image columns are ordered in the same order as recorded during seismic shot, so exists a strong spatial correlation between them. This way, machine learning techniques, such as those based on Convolutional Neural Networks (CNNs), can be valuable tools in accomplishing these tasks. Recently, CNNs have also been applied to other problems pertaining to seismic imaging, such as seismic texture classification~\cite{chevitarese2018efficient}, seismic facies classification~\cite{zhao2018seismic}, seismic fault detection~\cite{pochet2019seismic}, and salt segmentation~\cite{shi2019saltseg}.

%The abscissa represents the position of the sensor relative to the shot position

%Applications of ANNs in geophysics include seismic trace quality classification~\cite{mccormack1991neural}, first break signal recognition~\cite{ashida1996data}, and seismic window classification~\cite{diersen2011classification}. In seismological applications, ANNs are used in~\cite{valentine2010approaches} and \cite{paitz2017neural} to identify the quality of waveforms. In conventional processing, the identification of anomalous events is performed by analyzing the amplitude and frequency content of seismic traces ~\cite{elboth2009attenuation,bekara2014toward}.

In this letter, we focus on the \textbf{noise localization} process. Modern CNN-based frameworks for object detection, such as the Faster R-CNN~\cite{ren2015faster}, YOLO~\cite{redmon2018yolov3}, and single-shot multibox detector (SSD)~\cite{liu2016ssd}, focus on the recognition and localization of highly structured objects (e.g., cars, bicycles, and airplanes) or living entities (e.g., humans, dogs, and horses) rather than on unstructured scenes such as seismic shot gather noise. In this research, we investigate the effectiveness of a CNN-based detector for seismic shot gather noise localization. More precisely, We evaluate how a multi-scale feature-fusion-based neural network performs for for this task. We use a feature fusion approach because it integrates information from different feature maps with different receptive fields, allowing the finer layers to make use of the context learned from the coarser layers. Other studies \cite{zhang2019geospatial, zhu2018bidirectional, chen2019object} have experimented with fusing multi-scale feature layers of backbones, achieving considerable improvement as generic feature extractors in several applications in object detection and segmentation.

Our proposed detector is based on the following: 1) a feature-fusion-based backbone by combining MobileNet~\cite{howard2017mobilenets} and the Feature Pyramid Network (FPN)~\cite{lin2017feature}; 2) the SSD framework for noise detection on a multi-scale level; and 3) the focal loss function from RetinaNet~\cite{lin2017focal} to improve prediction accuracy. We constructed a real-world dataset containing 6,500 seismic shot gather images and 14,101 noise bounding boxes. Additionally, we conducted an experiment to demonstrate the contribution of each component to the proposed detector.

This letter is structured as follows. In Section II, we describe the construction of our dataset. In Section III, we present our proposed model, and in Section IV, we describe the experiment and provide an analysis of the results. Finally, in Section V, we present our conclusions and discuss future work.
\section{Seismic shot-gather dataset for noise localization}
\label{sec:dataset}

Our dataset is derived from an offshore towed in a targeted region with 7,993 shot gathers from eight cables each for a total of 63,944 shot gather images. Of the total generated images, 6,500 were randomly selected and manually classified by a geophysicist with ``Good" and ``Bad" labels based on the visual inspection of artifacts related to swell noise and anomalous recorded amplitude. This resulted in two sets with 1,579 and 4,921 images for the ``Good" and ``Bad" labels, respectively. Fig.~\ref{fig:seismic_classification} presents two examples of shot gather images classified by a geophysicist.

To accommodate the CNN input, we resized all shot gather images to a square format of 600 x 600 pixels. Then, geophysicists used the VoTT\footnote{https://github.com/Microsoft/VoTT} tool to manually annotate bounding boxes around the noise regions for each image in the ``Bad" set, resulting in 14,101 annotations (bounding boxes).

\begin{figure}
    \centering
    \includegraphics[width=0.4\textwidth]{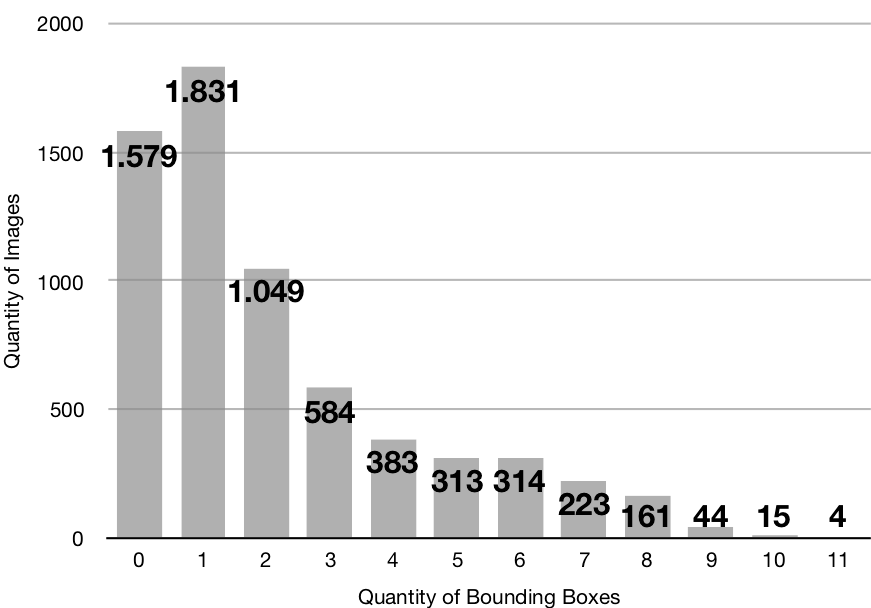}
    \caption{Shot gather image distribution per number of annotated bounding boxes. Column '0' represents the number of images in the ``Good'' set, while the other columns represent the number of bounding boxes in ``Bad'' set.}
    \label{fig:bb_histogram}
\end{figure}

Finally, we split our dataset into 80\%, 10\%, and 10\% for the training, validation, and testing sets, respectively. We balanced these sets by analyzing the dataset distribution, which is illustrated in Fig.~\ref{fig:bb_histogram}. Then, we proportionally selected images from each column to maintain a similar distribution for all three sets. Our final dataset had 5,200 images for training, 650 images for validation, and 650 images for testing.

\section{Noise Detection Network}
\label{sec:models}

A CNN-based detector is generally composed of two modules. The first is referred to by researchers as the \textit{backbone}, which acts as the feature extractor that provides the detector with discriminating power. The second module, the \textit{detector meta-architecture}, operates on the extracted features from the backbone to generate detection boxes. 

\begin{figure}[h]
    \centering
    \includegraphics[width=0.48\textwidth]{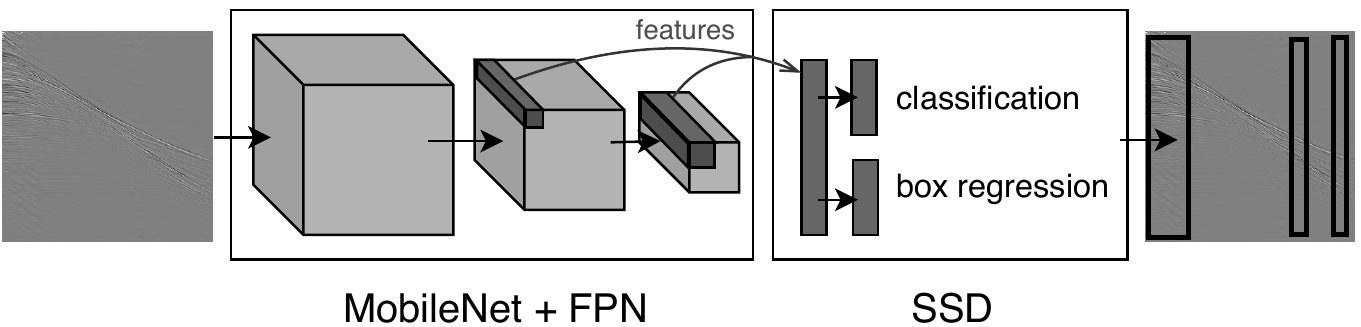}
    \caption{Overview of proposed network for noise localization in seismic shot-gather images.}
    \label{fig:proposed_net}
\end{figure}

Fig.~\ref{fig:proposed_net} illustrates our detector based on this structure. For the backbone, we use a feature fusion architecture that results from combining MobileNet with the FPN to generate convolutional features with rich semantic information on different scale levels. Our meta-architecture is based on the SSD structure, which performs the regression and classification of box coordinates over the features generated by the backbone. In addition, we use the focal loss rather than the classical cross-entropy-based loss function to improve the prediction accuracy of the SSD.   

The remainder of this Section is structured as follows. In  subsection~\ref{subsec:ssd_backbone}, we present the MobileNet+FPN backbone architecture, while in subsection~\ref{subsec:ssd}, we present the SSD framework. In subsection~\ref{subsec:focal_loss}, we present the focal loss function and its application.  

\subsection{MobileNet+FPN backbone}
\label{subsec:ssd_backbone}

Our backbone is based on a feature fusion model, combining MobileNet~\cite{howard2017mobilenets} and the  FPN~\cite{lin2017feature} into a single network that projects rich feature maps on three different scales. MobileNet is the core of the backbone. It is built on a  depth-wise separable convolution (3x3 depth-wise convolution, followed by 1x1 convolution) that requires 8 to 9 times less computation than traditional convolution. The FPN augments the MobileNet with lateral 1x1 convolutions and fuses feature maps of different scales by nearest-neighbor up-sampling and element-wise sum operation. The final output of the network consists of feature maps (called \textit{projections}) on three different scales; these are used by the SSD for noise detection. 

\begin{figure*}[h]
    \centering
    \includegraphics[width=\textwidth]{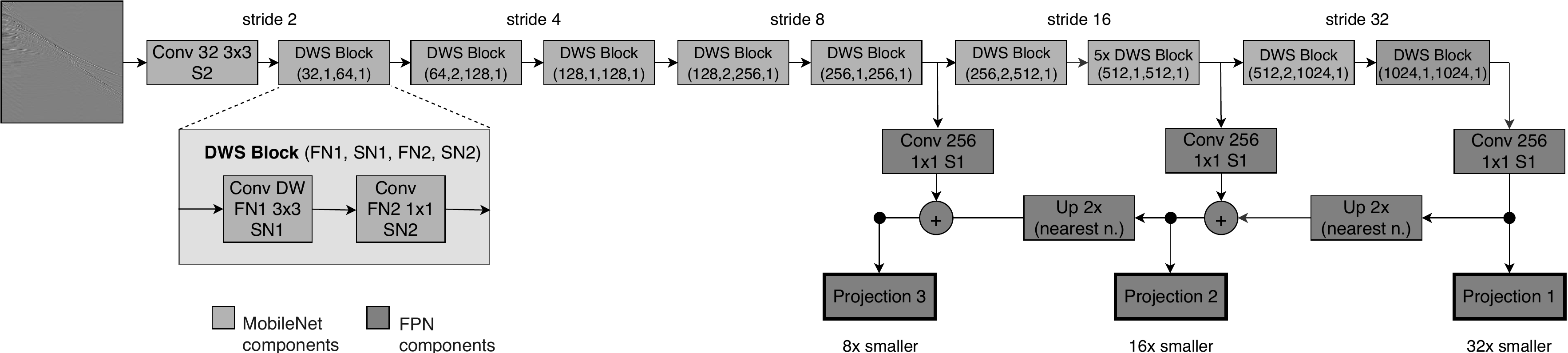}
    \caption{MobileNet+FPN Network}
    \label{fig:mob_fpn}
\end{figure*}

Fig.~\ref{fig:mob_fpn} depicts the MobileNet+FPN architecture. The notation ``Conv 32 3x3 S2" denotes a convolutional layer with 32 filters, a 3x3 kernel and stride 2. The MobileNet consists of a Conv layer with stride 2 followed by 13 depth-wise separable (DWS) blocks. Internally, each DWS block has a 3x3 depth-wise convolution followed by a 1x1 convolution (also called point-wise convolution). Then, the FPN  processes the lateral features maps from 8, 16 and 32 strides with a 1x1 convolution and combines them by element-wise summation after nearest-neighbor up-sampling. The final three projections used for noise detection are 32, 16 and 8 times smaller than the input image. All convolutional layers use batch normalization and ReLU activation. 

\subsection{Single Shot Multibox Detector}
\label{subsec:ssd}

The SSD~\cite{liu2016ssd} is a single stage framework for object detection with an accuracy similar to other state-of-the-art detectors such as YOLO and Faster R-CNN. Owing to the variance in noise size, we selected the SSD to take advantage of its multi-scale box matching strategy. The SSD operates by creating thousands of default boxes corresponding to different regions on three feature maps generated by the MobileNet+FPN backnone. The SSD determines which default boxes correspond to ground truth detection and trains the network accordingly. For each ground truth box, the SSD selects the most appropriate box from the default boxes by matching it to the default box with the best intersection over union (IoU) coefficient (higher than a threshold of 0.5). During training, the SSD learns to predict class scores (in our case, classes 0 and 1 for background and noise, respectively) and box offsets from the selected default boxes.

The SSD achieves its objective with the help of a multitask loss function, which is the weighted sum of the confidence loss (conf) and  localization loss (loc) as follows:
\begin{equation}
\label{equ:}
L(x,c,l,g) = \frac{1}{N}(L_{conf}(x,c) +{\alpha}L_{loc}(x,l,g))    
\end{equation}

where N is the number of matched default boxes. Let $x_{ij}^p = \{1,0\}$ be an indicator for matching the $i$-th default box to the $j$-th ground truth box of category $p$. The confidence loss is the \textit{softmax} loss over the confidence of multiple classes ($c$):
\begin{equation}
\begin{gathered}
L_{conf}(x,c) = - \sum_{i \in Pos}^{N} x_{ij}^p{log}(\hat{c}_i^p) - \sum_{e \in Neg}log(\hat{c}_i^0) \\
where \quad \hat{c}_i^p = \frac{exp(c_i^p)}{\sum_p{exp(c_i^p)}}
\end{gathered}
\end{equation}

The localization loss is a Smooth L1 loss between the predicted box $l$ and the offset box $\hat{g}$. The offset box is calculated from  ground truth box $g$ and default box $d$. The parameters $cx, cy$, $w$ and $h$ denotes the center, width and height of the box, respectively.
\begin{equation}
\begin{gathered}
L_{loc}(x,l,g) \sum_{i \in Pos}^{N} \sum_{m \in \{cx,cy,w,h\}} x_{ij}^p{smooth}_{L1}(l_i^m - \hat{g}_j^m) \\ 
where \quad {smooth}_{L1}(z) = \begin{cases} 0.5z^2 \quad if|z| < 1 \\ |z| - 0.5 \quad otherwise,\end{cases}\\
\hat{g}_j^{cx} = ({g}_j^{cx} - {d}_i^{cx})/{d}_i^{w} \quad \hat{g}_j^{cy} = ({g}_j^{cy} - {d}_i^{cy})/{d}_i^{h} \\
\hat{g}_j^{w} = log({g}_j^{w}/{d}_i^{w}) \quad \hat{g}_j^{h} = log({g}_j^{h}/{d}_i^{h})    
\end{gathered}
\end{equation}

\begin{figure}[h]
    \centering
    \includegraphics[width=0.5\textwidth]{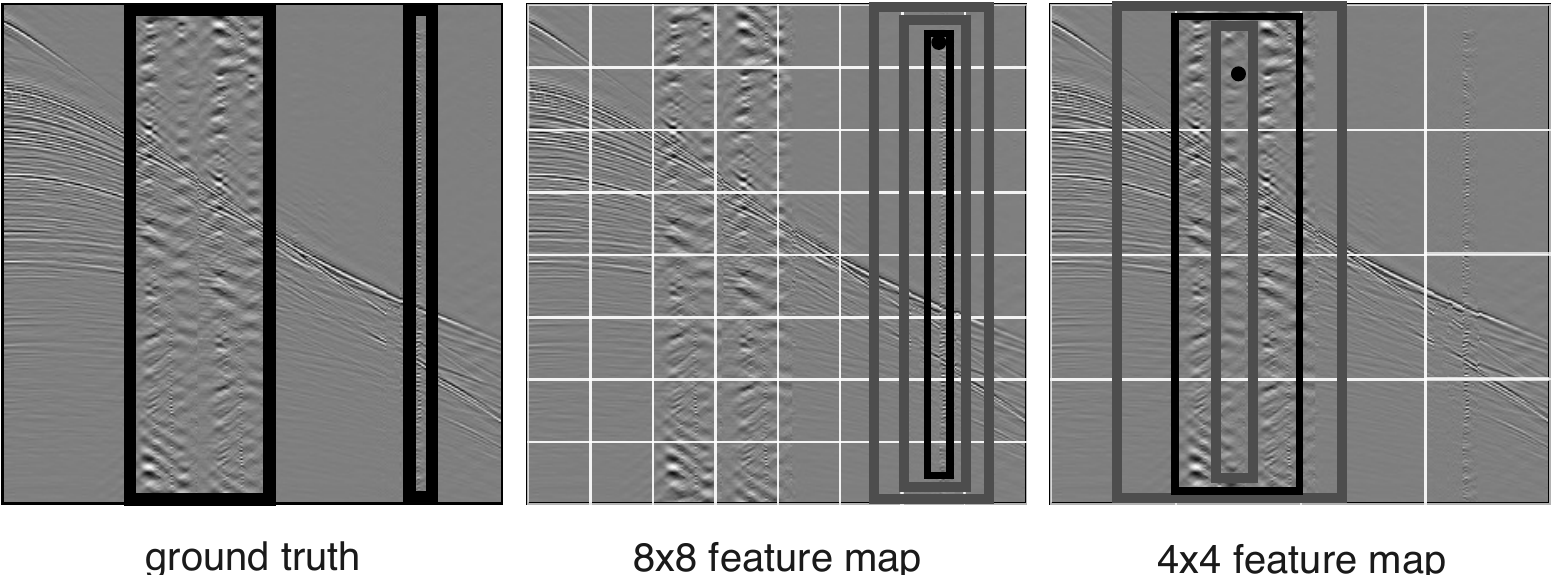}
    \caption{Example of box matching strategy over 8x8 and 4x4 feature maps.}
    \label{fig:sample_boundingbox}
\end{figure}

By combining predictions for all default boxes with different scales and aspect ratios from all positions of features maps generated by the backbone network, the SSD obtains a diverse set of predictions, covering various input noise sizes and shapes. Because many boxes attempt to localize objects during the inference step, a post-processing step called greedy non-maximum suppression (NMS) is applied to suppress duplicate detection. For example, as illustrated in Fig.~\ref{fig:sample_boundingbox}, given an input image with the ground truth, the large noise is matched to a default box in a position on the 4x4 feature map, while the thin noise is matched to another position on the 8x8 feature map. This occurs because the boxes have different scales and matching is performed in the most appropriate feature map.

\subsection{Focal Loss}
\label{subsec:focal_loss}

The focal loss function was introduced by RetinaNet~\cite{lin2017focal} and solves the foreground-background class imbalance problem in one-stage detectors. As described in Subsection \ref{subsec:ssd}, the SSD evaluates thousands of default boxes; however, most of these boxes do not contain noise (negative examples). The principle of the focal loss function is to reduce the load of these simple negative boxes in order for the loss to focus on boxes with useful content, which can improve the prediction accuracy.

We first introduce the cross-entropy loss (CE) for binary classification:
\begin{equation}
\begin{gathered}
CE(p,y) = \begin{cases} -log(p) \quad if\quad y = 1 \\ -log(1-p) \quad otherwise.\end{cases}\\
\end{gathered}
\end{equation}

In the above $y \in \{\pm 1\}$ specifies the ground truth class, while $p \in [0,1]$ is the model's estimated probability for the class with label $y=1$. For notational convenience, $p_t$ is defined as:
\begin{equation}
\begin{gathered}
p_t = \begin{cases} p \quad if\quad y = 1 \\ 1-p \quad otherwise,\end{cases}
\end{gathered}
\end{equation}
and then $CE(p,y) = CE(p_t) = -log(p_t)$.

The focal Loss, adds a modulating factor $(1 - p_t)^{\gamma}$ to the cross entropy function. The tunable focusing parameter $\gamma \ge 0$ reduces the relative loss for the simple examples. In this work, we use the $\alpha$-balanced variant of the focal loss, where weighting factor $\alpha \in [0,1]$ is used to balance the importance of negative/positive examples. For notational convenience, $\alpha_t$ is defined as: 
\begin{equation}
\begin{gathered}
\alpha_t = \begin{cases} \alpha \quad if\quad y = 1 \\ 1-\alpha \quad otherwise,\end{cases}\\
\end{gathered}
\end{equation}

The $\alpha$-balanced focal loss is defined as:
\begin{equation}
\begin{gathered}
FL(p_t) =-\alpha_t{(1 - p_t)}^{\gamma}log(p_t)
\end{gathered}
\end{equation}

%With this rescaling, the large number of easily classified examples (mostly background) does not dominate the loss anymore and learning can concentrate on the few interesting cases.
\section{Experimentation}
\label{sec:experiment}

In this section, we evaluate the effectiveness of the proposed approach for seismic shot-gather noise localization. First, to attest the choice of our basic backbone, we compare the results of MobileNet with the other  two popular CNNs, VGG16~\cite{simonyan2014very}, and InceptionV3~\cite{szegedy2016rethinking}. Next, we supplement the MobileNet with FPN and Focal Loss and evaluate each component to determine the contribution to the final architecture. To this end, we created three backbone models: 1) MobileNet+FPN; 2) MobileNet+FocalLoss; and 3), our proposed network, MobileNet+FPN+FocalLoss. 

%In addition, we compared the performance of each backbone by initializing it with and without the transfer learning from the COCO dataset\footnote{http://cocodataset.org}. 

%In Subsection~\ref{subsec:configuration} we describe the training configuration. Next, n Subsection~\ref{subsec:metrics}  we describe the evaluation metrics. And finally, in Subsection~\ref{subsec:results} we present our empirical findings.

We decided to use the average precision (AP)\footnote{http://cocodataset.org/{\#}detection-eval} metric, as it is a popular metric for measuring detector performance. In problems related to localization, the AP is calculated over an IoU threshold. We used two AP metrics: the traditional AP@0.5 and the AP@[0.5:0.05:0.95]. The latter corresponds to the average of 10 IoU thresholds from 0.5 to 0.95 with a step size of 0.05.

\subsection{Training configuration}
\label{subsec:configuration}

Our networks were trained using a octa-core i7 3.40GHz CPU with a GTX-1070Ti GPU. The training used RMSprop~\cite{tieleman2012lecture} optimization with a momentum of 0.9, a decay of 0.9 and epsilon of 0.1; batch normalization with a decay of 0.9997 and epsilon of 0.001; fixed learning rate of 0.004; L2 regularization with 4e-5 weight; focal loss with alpha of 0.7 and gamma of 2.0; and batch size of 32 images and 200 epochs for training. In the NMS step an IoU coefficient of 0.6 was used to suppress duplicate detection.

\subsection{Results}
\label{subsec:results}

As illustrated in Table~\ref{tab:exp1_basic}, the best result was achieved by the MobileNet, which produced an AP@0.5 of 72.11\% and AP@[0.5:0.05:0.95] of 32.80\% followed by the InceptionV3, which produced an AP@0.5 of 70.94\% and AP@[0.5:0.05:0.95] of 32.71\%. The worse result was produced by the VGG16. with an AP@0.5 of 66.04\% and AP@[0.5:0.05:0.95] of 29.89\%.

\begin{table}[h]
   \renewcommand*{\arraystretch}{1.4}
    \centering
    \caption{Results of basic backbone models on validation set}
    \begin{tabular}{|l|l|c|c|}
        \hline
         \cellcolor{gray!30}\textbf{\#} &
         \cellcolor{gray!30}\textbf{Backbone} &  \cellcolor{gray!30}\textbf{AP@0.5 (\%)} & \cellcolor{gray!30}\textbf{AP@[0.5:0.05:0.95] (\%)} \\
         \hline
         1 & VGG16 & 66.04 & 29.89 \\
         \hline
         2 & IncepionV3 & 70.94 & 32.71 \\
         \hline
         3 & \textbf{MobileNet} & \textbf{72.11} & \textbf{32.80} \\
         \hline
    \end{tabular}
    \label{tab:exp1_basic}
\end{table}

Considering MobileNet supplemented by FPN and Focal Loss scenario, architecture \#2 produced the best AP@0.5 of 78.90\%, while architecture \#3 produced the best AP@[0.5:0.05:0.95] of 45.62\%, as illustrated in Table~\ref{tab:exp1_without_transf}. Because the latter metric is stricter than the former, we consider the architecture \#3 to be the winner. Not only the two architectures that used the focal loss were the ones that produced the best performances but also the focal loss performed better in combination with the FPN. In fact, using the FPN without the focal loss was less effective approach.

\begin{table}[h]
   \renewcommand*{\arraystretch}{1.4}
    \centering
    \caption{Results of scenario with MobileNet supplemented by FPN and Focal Loss on validation set}
    \begin{tabular}{|l|l|c|c|}
         \hline
         \cellcolor{gray!30}\textbf{\#} &
         \cellcolor{gray!30}\textbf{Backbone} &  \cellcolor{gray!30}\textbf{AP@0.5 (\%)} & \cellcolor{gray!30}\textbf{AP@[0.5:0.05:0.95] (\%)} \\
          \hline
          \multirow{2}{*}{1} &  \multirow{2}{*}{\shortstack[l]{MobileNet \\+ FPN}} &  \multirow{2}{*}{74.71} &  \multirow{2}{*}{41.12} \\
         & & & \\
         \hline
          \multirow{2}{*}{2} & \multirow{2}{*}{\shortstack[l]{MobileNet \\+ Focal Loss}} & \multirow{2}{*}{\textbf{78.90}} &  \multirow{2}{*}{40.08} \\
          & & & \\
         \hline
          \multirow{2}{*}{3} & \multirow{2}{*}{\shortstack[l]{\textbf{MobileNet} \\ \textbf{+ FPN + Focal Loss}}} &  \multirow{2}{*}{78.37} & \multirow{2}{*}{\textbf{45.62}} \\
         & & & \\
         \hline
    \end{tabular}
    \label{tab:exp1_without_transf}
\end{table}

The results of the best model for the test set are presented in Table~\ref{tab:exp1_rest_results}. The MobileNet+FPN+FocalLoss network produced an AP@0.5 of 73.13\% and AP@[0.5:0.05:0.95] of 38.14\%. This shows that the test set is a little more complicated than the validation set, since there is a performance loss of 5.24\% in AP@0.5 and 7.48\% in AP@[0.5:0.05:0.95].

\begin{figure}[h]
    \centering
    \includegraphics[width=0.40\textwidth]{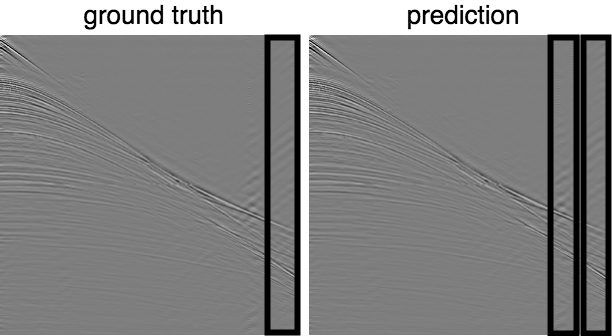}
    \caption{Prediction example of the MobileNet+FPN+FocalLoss on an example of test set. The predictor correctly localized two noises where just one was annotated by geophysicist}
    \label{fig:pred_examples}
\end{figure}

There are no other models using the bounding box matching strategy that specifically detect noise in seismic shot gather images. A comparison of our results with benchmark models for the generic object detection task (listed on the leader board of the COCO 2017 object detection task\footnote{\url{http://cocodataset.org/\#detection-leaderboard}}), reveals that our findings are similar to the scores of the benchmark models, since the first place on COCO ranking achieves an AP@0.5 of 73\%, thus indicating the effectiveness of our proposed model. Fig.~\ref{fig:pred_examples} presents an example prediction on the test set. Our predictor was more accurate than specialists; it correctly localized two noises, where only one had been annotated.

\begin{table}[h]
   \renewcommand*{\arraystretch}{1.4}
    \centering
    \caption{Results of the MobileNet+FPN+FocalLoss model on test set}
    \begin{tabular}{|l|c|c|}
         \hline
         \cellcolor{gray!30}\textbf{Backbone} &  \cellcolor{gray!30}\textbf{AP@0.5(\%)} & \cellcolor{gray!30}\textbf{AP@[0.5:0.05:0.95](\%)} \\
         \hline
         MobileNet+FPN+Focal Loss & 73.13 & 38.14 \\
         \hline
    \end{tabular}
    \label{tab:exp1_rest_results}
\end{table}

\section{Conclusion}
\label{sec:conclusion}

In this work, we investigated a multi-scale feature-fusion based neural network for noise localization in seismic shot gather images. We built a real-world dataset containing 6,500 seismic shot gather images and 14,101 bounding boxes of regions with noise. Our proposed detector model used MobileNet in combination with FPN as the backbone, an SSD as the detector meta-architecture, and focal loss. Our experiments revealed the contribution of each component of the proposed network. In the validation step, the proposed model achieved an AP@0.5 of 78.37\% and AP@[0.5:0.05:0.95] of 45.62\%. in the test step, it produced an AP@0.5 of 73.13\% and AP@[0.5:0.05:0.95] of 38.14\%.

To achieve higher performance, we plan to investigate the effectiveness of others mechanisms of state-of-the-art networks for object detection. Specifically, in  future work, we plan to extend our actual network with the ARM (Anchors Refinament Module) and ODM (Object Detection Module) of the RefineDet network~\cite{zhang2018single} aiming to further improve noise localization. Further future work involves the construction of a multitask network that both localizes and clears region with noise in seismic shot gather images.

\ifCLASSOPTIONcaptionsoff
  \newpage
\fi



%\bibliographystyle{IEEEtran}

\bibliographystyle{IEEEtran}
\bibliography{IEEEabrv,refs}

\begin{thebibliography}{10}
\providecommand{\url}[1]{#1}
\csname url@samestyle\endcsname
\providecommand{\newblock}{\relax}
\providecommand{\bibinfo}[2]{#2}
\providecommand{\BIBentrySTDinterwordspacing}{\spaceskip=0pt\relax}
\providecommand{\BIBentryALTinterwordstretchfactor}{4}
\providecommand{\BIBentryALTinterwordspacing}{\spaceskip=\fontdimen2\font plus
\BIBentryALTinterwordstretchfactor\fontdimen3\font minus
  \fontdimen4\font\relax}
\providecommand{\BIBforeignlanguage}[2]{{%
\expandafter\ifx\csname l@#1\endcsname\relax
\typeout{** WARNING: IEEEtran.bst: No hyphenation pattern has been}%
\typeout{** loaded for the language `#1'. Using the pattern for}%
\typeout{** the default language instead.}%
\else
\language=\csname l@#1\endcsname
\fi
#2}}
\providecommand{\BIBdecl}{\relax}
\BIBdecl

\bibitem{duarte2014seismic}
L.~T. Duarte, D.~Donno, R.~R. Lopes, and J.~M.~T. Romano, ``Seismic signal
  processing: Some recent advances,'' in \emph{2014 IEEE International
  Conference on Acoustics, Speech and Signal Processing (ICASSP)}.\hskip 1em
  plus 0.5em minus 0.4em\relax IEEE, 2014, pp. 2362--2366.

\bibitem{yilmaz2001seismic}
{\"O}.~Yilmaz, \emph{Seismic data analysis: Processing, inversion, and
  interpretation of seismic data}.\hskip 1em plus 0.5em minus 0.4em\relax
  Society of exploration geophysicists, 2001.

\bibitem{zhang2017beyond}
K.~Zhang, W.~Zuo, Y.~Chen, D.~Meng, and L.~Zhang, ``Beyond a gaussian denoiser:
  Residual learning of deep cnn for image denoising,'' \emph{IEEE Transactions
  on Image Processing}, vol.~26, no.~7, pp. 3142--3155, 2017.

\bibitem{elboth2009attenuation}
T.~Elboth, F.~Geoteam, and D.~Hermansen, ``Attenuation of noise in marine
  seismic data,'' in \emph{SEG Technical Program Expanded Abstracts
  2009}.\hskip 1em plus 0.5em minus 0.4em\relax Society of Exploration
  Geophysicists, 2009, pp. 3312--3316.

\bibitem{chevitarese2018efficient}
D.~S. Chevitarese, D.~Szwarcman, E.~V. Brazil, and B.~Zadrozny, ``Efficient
  classification of seismic textures,'' in \emph{2018 International Joint
  Conference on Neural Networks (IJCNN)}.\hskip 1em plus 0.5em minus
  0.4em\relax IEEE, 2018, pp. 1--8.

\bibitem{zhao2018seismic}
T.~Zhao, ``Seismic facies classification using different deep convolutional
  neural networks,'' in \emph{SEG Technical Program Expanded Abstracts
  2018}.\hskip 1em plus 0.5em minus 0.4em\relax Society of Exploration
  Geophysicists, 2018, pp. 2046--2050.

\bibitem{pochet2019seismic}
A.~Pochet, P.~H. Diniz, H.~Lopes, and M.~Gattass, ``Seismic fault detection
  using convolutional neural networks trained on synthetic poststacked
  amplitude maps,'' \emph{IEEE Geoscience and Remote Sensing Letters}, vol.~16,
  no.~3, pp. 352--356, 2019.

\bibitem{shi2019saltseg}
Y.~Shi, X.~Wu, and S.~Fomel, ``Saltseg: Automatic 3d salt segmentation using a
  deep convolutional neural network,'' \emph{Interpretation}, vol.~7, no.~3,
  pp. 1--36, 2019.

\bibitem{ren2015faster}
S.~Ren, K.~He, R.~Girshick, and J.~Sun, ``Faster r-cnn: Towards real-time
  object detection with region proposal networks,'' in \emph{Advances in neural
  information processing systems}, 2015, pp. 91--99.

\bibitem{redmon2018yolov3}
J.~Redmon and A.~Farhadi, ``Yolov3: An incremental improvement,'' \emph{arXiv
  preprint arXiv:1804.02767}, 2018.

\bibitem{liu2016ssd}
W.~Liu, D.~Anguelov, D.~Erhan, C.~Szegedy, S.~Reed, C.-Y. Fu, and A.~C. Berg,
  ``Ssd: Single shot multibox detector,'' in \emph{European conference on
  computer vision}.\hskip 1em plus 0.5em minus 0.4em\relax Springer, 2016, pp.
  21--37.

\bibitem{zhang2019geospatial}
X.~Zhang, K.~Zhu, G.~Chen, X.~Tan, L.~Zhang, F.~Dai, P.~Liao, and Y.~Gong,
  ``Geospatial object detection on high resolution remote sensing imagery based
  on double multi-scale feature pyramid network,'' \emph{Remote Sensing},
  vol.~11, no.~7, p. 755, 2019.

\bibitem{zhu2018bidirectional}
L.~Zhu, Z.~Deng, X.~Hu, C.-W. Fu, X.~Xu, J.~Qin, and P.-A. Heng,
  ``Bidirectional feature pyramid network with recurrent attention residual
  modules for shadow detection,'' in \emph{Proceedings of the European
  Conference on Computer Vision (ECCV)}, 2018, pp. 121--136.

\bibitem{chen2019object}
C.~Chen, W.~Gong, Y.~Chen, and W.~Li, ``Object detection in remote sensing
  images based on a scene-contextual feature pyramid network,'' \emph{Remote
  Sensing}, vol.~11, no.~3, p. 339, 2019.

\bibitem{howard2017mobilenets}
A.~G. Howard, M.~Zhu, B.~Chen, D.~Kalenichenko, W.~Wang, T.~Weyand,
  M.~Andreetto, and H.~Adam, ``Mobilenets: Efficient convolutional neural
  networks for mobile vision applications,'' \emph{arXiv preprint
  arXiv:1704.04861}, 2017.

\bibitem{lin2017feature}
T.-Y. Lin, P.~Doll{\'a}r, R.~Girshick, K.~He, B.~Hariharan, and S.~Belongie,
  ``Feature pyramid networks for object detection,'' in \emph{Proceedings of
  the IEEE Conference on Computer Vision and Pattern Recognition}, 2017, pp.
  2117--2125.

\bibitem{lin2017focal}
T.-Y. Lin, P.~Goyal, R.~Girshick, K.~He, and P.~Doll{\'a}r, ``Focal loss for
  dense object detection,'' in \emph{Proceedings of the IEEE international
  conference on computer vision}, 2017, pp. 2980--2988.

\bibitem{simonyan2014very}
K.~Simonyan and A.~Zisserman, ``Very deep convolutional networks for
  large-scale image recognition,'' \emph{arXiv preprint arXiv:1409.1556}, 2014.

\bibitem{szegedy2016rethinking}
C.~Szegedy, V.~Vanhoucke, S.~Ioffe, J.~Shlens, and Z.~Wojna, ``Rethinking the
  inception architecture for computer vision,'' in \emph{Proceedings of the
  IEEE conference on computer vision and pattern recognition}, 2016, pp.
  2818--2826.

\bibitem{tieleman2012lecture}
T.~Tieleman and G.~Hinton, ``Lecture 6.5-rmsprop: Divide the gradient by a
  running average of its recent magnitude,'' \emph{COURSERA: Neural networks
  for machine learning}, vol.~4, no.~2, pp. 26--31, 2012.

\bibitem{zhang2018single}
S.~Zhang, L.~Wen, X.~Bian, Z.~Lei, and S.~Z. Li, ``Single-shot refinement
  neural network for object detection,'' in \emph{CVPR}, 2018.

\end{thebibliography}

\end{document}